\documentclass{article}

\usepackage{times}
\usepackage{graphicx} 
\usepackage{subfigure}
\usepackage{pgfplots}

\usepackage{tikz}

\usepackage{natbib}

\usepackage{algorithm}
\usepackage{algorithmic}

\usepackage{hyperref}

\usepackage{tikz}

\usepackage[accepted]{icml2017} 

\icmltitlerunning{Actively Learning what makes a Discrete Sequence Valid}

\usepackage{amsmath}
\usepackage{amssymb}
\newcommand{\data}{\mathcal{X}}

\newcommand{\lik}{\mathcal{L}}

\newcommand{\set}[1]{\{#1\}}
\newcommand{\weights}{\mathcal{W}}

\newcommand{\out}{\mathrm{o}_t(x_{1:t-1}\mid\weights)}

\newcommand{\h}[1]{\mathrm{H}\!\left[#1\right]}
\newcommand{\st}{\,|\,}
\newcommand{\s}{\mathrm{s}}

\DeclareMathOperator*{\argmax}{argmax}

\pgfplotsset{
	physics/.style = {
		xtick pos=left,
		ytick pos=left,
		enlarge x limits=false,
		every x tick/.style={color=black, thin},
		every y tick/.style={color=black, thin},
		tick align=outside,
		xlabel near ticks,
		ylabel near ticks,
		axis on top,
	}
}

\makeatletter
\def\pgfplots@drawticklines@INSTALLCLIP@onorientedsurf#1{}%
\makeatother

\begin{document}

\twocolumn[
\icmltitle{Actively Learning what makes a Discrete Sequence Valid}

\icmlsetsymbol{equal}{*}

\begin{icmlauthorlist}
\icmlauthor{David Janz}{equal,cam}
\icmlauthor{Jos van der Westhuizen}{equal,cam}
\icmlauthor{Jos{\'e} Miguel Hern{\'a}ndez-Lobato}{equal,cam}
\end{icmlauthorlist}

\icmlaffiliation{cam}{University of Cambridge, UK}

\icmlcorrespondingauthor{David Janz}{dj343@cam.ac.uk}
\icmlcorrespondingauthor{Jos van der Westhuizen}{jv365@cam.ac.uk}

\icmlkeywords{active learning, recurrent neural network, sequences}

\vskip 0.3in
]

\printAffiliationsAndNotice{\icmlEqualContribution}

\begin{abstract}
Deep learning techniques have been hugely successful for
traditional supervised and unsupervised machine learning problems. In large
part, these techniques solve continuous optimization problems. Recently however,
discrete generative deep learning models have been successfully used to
efficiently search high-dimensional discrete spaces. These methods work by
representing discrete objects as sequences, for which powerful sequence-based
deep models can be employed. Unfortunately, these techniques are significantly
hindered by the fact that these generative models often produce invalid
sequences. As a step towards solving this problem, we propose to learn a deep
recurrent \textit{validator} model. Given a partial sequence, our model learns
the probability of that sequence occurring as the beginning of a full valid
sequence. Thus this identifies valid versus invalid sequences and crucially it also provides insight about how individual sequence elements influence
the validity of discrete objects. To learn this model we propose an approach
inspired by seminal work in Bayesian active learning. On a synthetic dataset, we
demonstrate the ability of our model to distinguish valid and invalid sequences.
We believe this is a key step toward learning generative models that faithfully
produce valid discrete objects.
\end{abstract}

\section{Introduction and Related Work} Generative models have seen many
fascinating developments in recent years such as the ability to produce realistic
images from noise \citep{radford_unsupervised_2015} and create artwork
\citep{gatys2016image}. One of the most exciting research directions in
generative modeling is using such models to efficiently search high-dimensional
discrete spaces \citep{gomez-bombarelli_automatic_2016,kusner17}. Indeed, discrete
search is at the heart of problems in drug discovery \citep{gomez2016design},
natural language processing \citep{Bowman2016,Guimaraes2017}, and symbolic
regression \citep{kusner17}.

Current methods for attacking these discrete search problems work by `lifting'
the search from discrete space to continuous space, via an
\emph{autoencoder} \cite{rumelhart1985learning}. Specifically, an autoencoder
jointly learns two mappings: 1) a mapping from discrete space to continuous
space called an \emph{encoder}; and 2) a reverse mapping from continuous space back
to discrete space called a \emph{decoder}. These mappings are learned so that if
we map a discrete object to a continuous one via the encoder, then map it back
via the decoder, we reconstruct the original object. The hope is that, once the
autoencoder is fully trained, the continuous space (often called the `latent'
space) acts as proxy for the discrete space. If this holds, we can use the
geometry of the continuous space to improve search using (Euclidean) distance
measures and gradients, among many other things.
\citet{gomez-bombarelli_automatic_2016} showed that is possible to use this
technique to search for promising drug molecules.

Unfortunately, these methods are severely hindered by the fact that the decoder
often produces invalid discrete objects. This happens because it is difficult to enforce valid syntax and semantics in the latent and discrete space. Powerful sequential models (e.g., LSTMs \citep{hochreiter1997long} GRUs \citep{cho2014learning},
DCNNs \citep{kalchbrenner2014convolutional}) can exploit the relationship
between parts of the discrete objects (e.g., comparing similar sequences of
atoms in different molecules). When employing these models as encoders and
decoders, generation of invalid sequences is still possible, and currently this happens frequently (see Table 6 in the
Supplementary Material of \citet{kusner17}). A recent method \citep{kusner17}
aimed to fix this by using a grammar to rule out generating certain invalid
sequences.  However, the grammar only describes syntactic constraints and cannot
enforce semantic constraints.  Therefore, certain invalid sequences can still
be generated using that approach.

In this work-in-progress paper, we propose a method for learning the
probability that a partial discrete sequence leads to a full valid sequence.
The motivation for this is that, given knowledge about the probability of
partial sequences, we can influence discrete generative models to only produce
sequences that, \emph{at any point}, have a high likelihood of being valid. We
propose learning a Bayesian Recurrent Neural Network
\citep{gal_theoretically_2015,fortunato2017bayesian} to approximate these
probabilities, given access to a function that labels full sequences as
valid/invalid (such functions already exist for molecules, symbolic
expressions, and many natural language processing problems). Unfortunately, as
the length of sequences grows, as well as the number of possible elements, it
quickly becomes impossible to observe all possible sequences during training.
Thus, we design a Bayesian active learning approach for training our model,
inspired by classic mutual-information-based approaches
\citep{houlsby2011bayesian,hernandez2014predictive}. We illustrate the accuracy
and efficiency of our approach on the task of learning to identify the
probability that partial subsequences will lead to syntactically and
semantically valid mathematical expressions in Python.

\section{Learning Validity}

We denote the set of discrete sequences of length $T$ as $\data =
\set{(x_1, \dots, x_T) \st x_t \in \set{1, \dots, C}}$, with an alphabet of size
$C$ and elements $\mathbf{x} = x_{1:T} \in \data$. We
assume the availability of a \textit{validator} $\s\colon \data \to \set{0, 1}$
specifying whether a given sequence is valid. Here, it is important to note
that such a validator gives very sparse feedback: a sequence can only be
labelled \textit{as is}. Complete full-length sequences are thus assigned
meaningful labels, but arbitrary sub-sequences, whilst capable of making up a
valid sequence, may not be labelled. We aim to construct a model
for the probability of a sequence compiling at intermediate steps $t$ of its
generation, $\mathbb{P}(\s(\mathbf{x}) = 1 \mid x_{1:t})$, such that it may be
used to guide the training of models capable of generating valid elements of
$\data$ with high probability.

For small problems, where the alphabet size and sequence length are short, this
probability mass function can be found exactly by enumeration. We focus on the
case where this is not a feasible, and instead train a recurrent neural network
to approximate these probabilities. The output of said neural network at each
time step $t$, conditioned on some weights $\weights$, is denoted $\out$, and
is a vector of probit outputs, one for each character in the alphabet. That is
\begin{equation}
  \label{eq:2} \out \big|_{k} \in [0,1], \quad k \in \set{1, \dots, C}\,,
\end{equation} When this recurrent neural network is sufficiently flexible and it is combined with the cross-entropy-based loss
function
\begin{align}
  \label{eq:1} \lik(\weights \mid \mathbf{x}, s) & = \sum_{t=1}^T \s(\mathbf{x}) \log
\left[ \out\big|_{x_{t}} \right] + \nonumber\\ & \quad (1 - \s(\mathbf{x})) \log \left[ 1 -
\out\big|_{x_{t}} \right]\,,
\end{align}
we obtain that, when training is done by sampling
sequences uniformly from $\data$,
the value of $\weights$ that
minimizes the loss function satisfies $\out\big|_{x_t} = \mathbb{P}(\s(\mathbf{x}) = 1 \mid x_{1:t})$. Therefore, after such training process, the
resulting model can be used to generate sequences that are valid. For example,
by sampling at step $t$ only those values of $x_t$ from $\out\big|_{x_t} =
\mathbb{P}(\s(\mathbf{x}) = 1 \mid x_{1:t})$. The proposed model is end-to-end differentiable, meaning that stochastic
gradient descent can be used to find a local minimum of (\ref{eq:1}).

\section{Efficient Active Learning}

In general, for successful training, a reasonable
fraction of the samples seen by a model need to be valid. However, typically, as the
length of the sequences in $\mathcal{X}$ increases, the fraction of sequences
containing no errors, and thus being valid, tends to zero. To learn
effectively in this scenario, instead of sampling data from $\data$ uniformly,
we pursue an information-theoretic active learning strategy
\cite{mackay1992information}. In particular, we iteratively construct our
training set by sampling at sequences $\mathbf{x}\in \data$
that approximately maximise the gain of information on $\weights$ when
$\s(\mathbf{x})$ is observed. This will reduce the amount of data that is
needed in order to identify $\weights$.

To implement the active learning strategy described by
\citet{mackay1992information}, we need to follow a Bayesian approach. For this,
we use recent work by \citealt{gal2016theoretically} linking dropout and
approximate Bayesian inference. Under this approach, sampling from the neural
network's posterior distribution over the weights is approximated by applying a
dropout mask. Note that other approaches for approximate Bayesian inference in
recurrent neural netowkrs also exist \cite{fortunato2017bayesian} and could
have been used as well.

\paragraph{Maximising the information gain}

The expected gain of information $J(\mathbf{x})$ obtained by incorporating $\{
\mathbf{x}, \s(\mathbf{x}) \}$ into the training data can be measured in terms
of the expected reduction in entropy of the posterior distribution for
$\weights$. The quantity $J(\mathbf{x})$ is equivalent to the mutual
information between $\weights$ and the labels generated by the model, that is,
the $T$ binary variables assumed to be sampled from Bernoulli distributions
with probabilities $\out\big|_{x_t}$ for $t=1,\ldots,T$, where $\weights$ is
sampled from the model's posterior distribution
\cite{houlsby2011bayesian,NIPS2012_4700}. Optimizing $J(\mathbf{x})$ with
respect to $\mathbf{x}$ is infeasible in practice.
Instead, we follow a greedy approach and we iteratively select the $t$-th character $x_t$ in
$\mathbf{x}$ by optimizing the mutual information between
the $t$-th binary variable sampled from the model and $\weights$ when the input
to the model is $x_{1:t-1}$. In particular, we optimize
\begin{align}
J(x_t \mid x_{1:{t-1}}) & =
\h{\mathbb{E}_{\weights \mid \mathcal{D}}\:\out\big|_{x_t}} - \nonumber\\
& \quad \quad \quad \quad \,\, \mathbb{E}_{\weights \mid \mathcal{D}}\:\h{\out\big|_{x_t}}\,.\label{eq:infogain}
\end{align}
The expectations in this expression can be approximated by Monte Carlo, by
repetitively applying a random dropout mask and computing the network's output.
The efficacy of dropout-based Bayesian neural networks for active learning has
previously been established in the context of image data by
\citealt{gal2016active}. We denote by $\tilde{\mathbf{x}}^\star$ the sequence obtained
by iteratively optimizing $J(x_t \mid x_{1:{t-1}})$ from $t=1$ to $t=T$.

\paragraph{Generating minibatches}

The previous approach generates a single most informative sequence
$\tilde{\mathbf{x}}^\star$. However, in practice, we would like to generate a
minibatch of sequences to efficiently update the model by processing multiple
data points at a time. One possibility for this would be to repeatedly apply
the previous approach to sample mutiple $\tilde{\mathbf{x}}^\star$. However, this would result
in a collection of sequences that are individually informative but not diverse.
To introduce diversity and enforce exploration, we propose to add artificial
noise to the expectations in (\ref{eq:infogain}) during the construction of
$\tilde{\mathbf{x}}^\star$. The amount of injected noise then determines the
diversity of the elements in the resulting minibatch
$\set{\tilde{\mathbf{x}}^\star_n \: \mid \: n=1, \dots, N}$.

A simple way to introduce noise in (\ref{eq:infogain}) is to use a
small number of samples when approximating the expectations by Monte Carlo.
This is akin to Thompson sampling (TS), commonly used in Bayesian optimisation
\cite{hernandez2017parallel} and reinforcement learning
\cite{chapelle2011empirical} to balance exploration and exploitation.
TS aims at selecting the next input that optimizes the expected
value of the objective given the current data. However, this is a purely
exploitative criteron. To introduce exploration, TS
approximates the expectation over objective functions with a single sample from
the posterior distribution. However, unlike TS, here we have to
utilise more than a single sample to approximate our expectations, as using a
single sample gives always $J(x_t \mid x_{1:t-1}) = 0$ for any $x_t$. In our
experiments, we use a two sample estimate.
The corresponding pseudocode is shown in Algorithm \ref{alg:main}.

In our experiments we found that convergence was improved by starting the
training process with a small amount of data sampled from $\mathcal{X}$
uniformly at random rather than by using the proposed active learning method.
This tendency of active learning methods to perform poorly when they are
applied too early during training has already been reported before
\cite{seeger2008bayesian}, and in in our it is likely due to the uncertainty
estimates not being well calibrated when the model is randomly initialised.

\begin{algorithm}[hb]
   \caption{Generating a minibatch using active learning}
   \label{alg:main}
\begin{algorithmic}[1] 
\small

  \FOR{$t=1$ {\bfseries to} $T$}
  \FOR{$n=1$ {\bfseries to} $N$}

  \STATE sample $\weights_{n,k} \sim p(\weights  \mid  \mathcal{D})$ for $k=1,2$
  \STATE run forward pass, obtaining $\mathrm{o}_{t,n,k}(x_{1:t-1,n} \mid \weights_{n,k})$
  \STATE $\tilde{x}^\star_{t,n} \leftarrow \argmax_{x_t\in\set{1,\dots,C}} J(x_{t,n} \mid x_{1:t-1,n})$ where
  the expectations are taken across samples for $k=1,2$
  \ENDFOR
  \ENDFOR
  \STATE
return $\set{\mathbf{\tilde{x}}^\star_n \: \mid \: n=1, \dots, N }$
\end{algorithmic}
\end{algorithm}

\section{Results}

As a preliminary test for the proposed method, we consider learning the validity
of simple mathematical expressions. We define an alphabet
consisting of integers and some of the mathematical symbols allowable in the Python
programming language: \texttt{0123456789-*+/=<>()!}. Sequences generated from
this alphabet are executed as Python code to test their validity. In
particular, $\s(\mathbf{x}=0)$ if the sequence causes an error, else  $\s(\mathbf{x}) =
1$. Example errors include
\texttt{SyntaxError} if the sequence cannot be parsed into a valid Python parse
tree and runtime errors such as \texttt{OverflowError} and
\texttt{ZeroDivisionError}.

To test our method, we train three recurrent neural networks with dropout, all
of them identical  except for the data that they
are trained on.
The models are LSTMs with one hidden layer and 100 hidden units.
The networks receive sequences of length $T=25$ and have the following characteristics:

\begin{itemize}
\item \textit{Vanilla}: uses sequences sampled uniformly from $\data$. These
contain approximately 0.1\% positive examples.

\item \textit{Balanced}: samples uniformly from
$\data$, rejecting negative examples until at least 2\% of the samples are positive.
\item \textit{Active}: uses active learning as described in this paper to
generate training minibatches.
\end{itemize}

\begin{figure*}
\centering
\scalebox{0.8}{
		\begin{tikzpicture}
		\begin{axis}[ physics, xlabel={training points seen},
ylabel={average AUC}, legend pos= south east, xmax = 150000, ymax = 1.0 ]
\addplot+ [no marks, thick] table [x=Step, y=avg_value_3, col sep=comma,
mark=none] {fig_data/vanilla.csv}; \addlegendentry{Vanilla}
		\addplot+ [no marks, thick] table [x=Step, y=avg_value_3, col
sep=comma, mark=none] {fig_data/balanced.csv}; \addlegendentry{Balanced}

		\addplot+ [no marks, thick] table [x=Step, y=avg_value_3, col
sep=comma, mark=none] {fig_data/active.csv}; \addlegendentry{Active}
		\end{axis}
		\end{tikzpicture}
}
\hspace{0.25cm}
\scalebox{0.8}{
		\begin{tikzpicture}
		\begin{axis}[ physics, xlabel={training time/seconds},
ylabel={average AUC}, legend pos= south east, xmax=350, ymax=1.0, ] \addplot+
[no marks, thick] table [x=Relative_time, y=avg_value_3, col sep=comma,
mark=none] {fig_data/vanilla.csv}; \addlegendentry{Vanilla}

		\addplot+ [no marks, thick] table [x=Relative_time,
y=avg_value_3, col sep=comma, mark=none] {fig_data/balanced.csv};
\addlegendentry{Balanced}

		\addplot+ [no marks, thick] table [x=Relative_time,
y=avg_value_3, col sep=comma, mark=none] {fig_data/active.csv};
\addlegendentry{Active}
		\end{axis}
		\end{tikzpicture}
}
\caption{Left, comparison in performance of the three neural networks as a
function of the number of data points they have seen in training.
Right, comparison in performance of the three neural networks as a
function of time spent training and generating training minibatches.}\label{fig:relative}
\end{figure*}
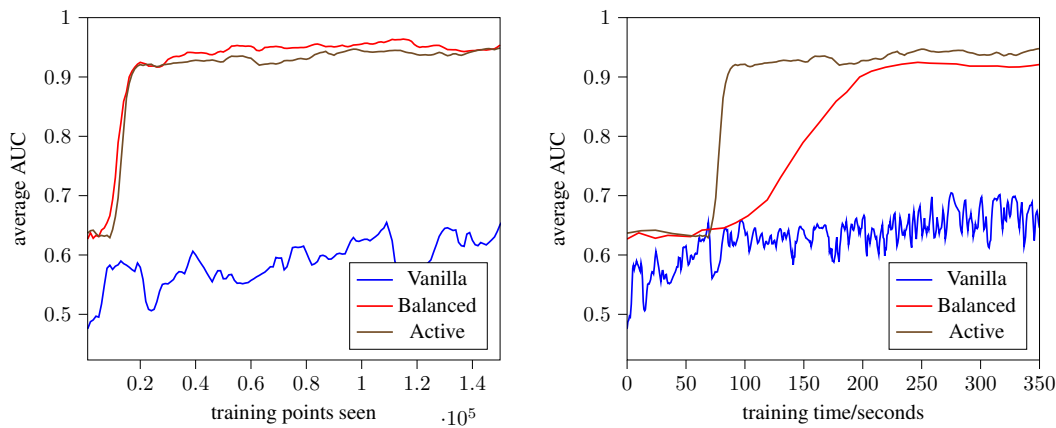

The performance of each network is evaluated on a validation set that is fully
balanced -- containing 50\% positive and 50\% negative examples. As a validation
metric we use the area under the receiver operating characteristic curve (AUC),
commonly used when working with binary classifiers. To assess the networks'
ability to make predictions on partial sequences, the AUC quantity reported is
the average AUC obtained when considering each of the $T$ subsequences forming
each sequence, where the subsequences start at $t=1$ and finish at
$t=1,\ldots,T$.
The AUC can be understood as the probability that a uniformly drawn positive
example is ranked before a uniformly drawn negative example.
The reason for using the AUC metric is that it
is largely insensitive to changes in the proportion of positive examples seen during
training, which will be affected by the sampling strategy used.

The plot in the left part of Figure \ref{fig:relative} shows the average AUC
metric obtained by each of the methods as a function of the number of sequences seen thus far in training.
This plot shows that the \textit{Vanilla}-based model fails to learn an accurate
ranking of sequences. This is because this method samples training sequences
containing far too few positive samples. By contrast, the methods
\textit{Balanced} and \textit{Active} perform much better than \textit{Vanilla}
since they use a more balanced set of training sequencces.

The right hand side plot in Figure \ref{fig:relative} shows again the average AUC values but now as a
function of the wall-clock time used by each method. Here we can see that
\textit{Balanced} is computationally very expensive, even for the simple
scenario considered here, and that generating sequences actively proves much
more efficient.  We expect that for real sequences, such as strings encoding
molecular structures, and for longer sequences, the differences in performance
between \textit{Active} and \textit{Balanced} will be larger --
\textit{Balanced} will quickly become computationally infeasible.

\section{Sampling Sequences from the Models}

We now consider how to sample valid sequences from the learned models.  For
this, we follow Boltzmann sampling technique. At each step, we sample from a Boltzmann distribution with some temperature $\theta$ and weights given by the output of the neural network for each $x_t$ in the alphabet. The temperature parameter allows us to  trade-off between sequence diversity and uniqueness. At best, \textit{Vanilla} produces 10\% valid sequences; both \textit{Balanced} and \textit{Active} can achieve 100\% validity at similar levels of diversity.

\section{Discussion and Future Work}

In this work-in-progress paper we have shown how to efficiently learn a
conditional model for the validity of sequences, and demonstrated its
effectiveness on an example consisting of arithmetic expressions. Our approach
is based on tractable approximations to information-theoretic active learning
and it allows us to handle hugely unbalanced data in a principled manner.

The proposed conditional model can predict the correctness of an expression
part way through the generation process. This allows for such model to be used
as a guide during the training of an autoencoder, by biasing the decoder
towards valid sequences, thus making its learning process easier.  We leave the
exact details of combining the proposed conditional model with a generative
model such as an autoencoder for future work; for one possible approach see
\cite{kusner17}.

In follow up work we plan to evaluate the proposed approach when working with
sequences ecoding molecular structures. Building accurate generative models for
valid molecular structures has important applications in chemical design. The
existence of benchmarks in this domain should allow us to show quantifiably
superior results over other existing approaches.  We shall also investigate
more principled ways of injecting noise into the entropy estimates used for
generating training minibatches, consider other types of Bayesian recurrent
neural networks and explore possible methods for encoding more structural
information a priori by, for example, by operating on parse trees rather than
on a character-by-character basis.

\bibliography{bgrnn,jos_library}
\bibliographystyle{icml2017}

\end{document}